\documentclass{article}
\usepackage{ijcai16}

\usepackage{times}
\usepackage{epsfig}
\usepackage{graphicx}
\usepackage{xcolor}

\title{Action Recognition Based on Joint Trajectory Maps Using\\ Convolutional Neural Networks}

\author{Pichao Wang$^{\rm 1}$\thanks{Both authors contributed equally to this work}, Zhaoyang Li$^{\rm 2}$\footnotemark[1], Yonghong Hou$^{\rm 2}$\thanks{Corresponding author} and Wanqing Li$^{\rm 1}$\\
$^{\rm 1}$Advanced Multimedia Research Lab, University of Wollongong, Australia\\
$^{\rm 2}$School of Electronic Information Engineering, Tianjin University, China\\
{\tt\small pw212@uowmail.edu.au,lizhaoyang@tju.edu.cn, houroy@tju.edu.cn, wanqing@uow.edu.au}\\
}

\begin{document}

\maketitle

\begin{abstract}

Recently, Convolutional Neural Networks (ConvNets) have shown promising performances in many computer vision tasks, especially image-based recognition. How to effectively use ConvNets for video-based recognition is still an open problem. In this paper, we propose a compact, effective yet simple method to encode spatio-temporal information carried in $3D$ skeleton sequences into multiple $2D$ images, referred to as Joint Trajectory Maps (JTM), and ConvNets are adopted to exploit the discriminative features for real-time human action recognition. The proposed method has been evaluated on three public benchmarks, i.e., MSRC-12 Kinect gesture dataset (MSRC-12), G3D dataset and UTD multimodal human action dataset (UTD-MHAD) and achieved the state-of-the-art results.

\end{abstract}

\section{Introduction}
Recognition of human actions from RGB-D (Red, Green, Blue and Depth) data has attracted increasing attention in multimedia signal processing in recent years due to the advantages of depth information over conventional RGB video, e.g. being insensitive to illumination changes. Since the first work of such a type~\cite{li2010action} reported in 2010, many methods~\cite{wang2012mining,Oreifej2013,yangsuper,lurange} have been proposed based on specific hand-crafted feature descriptors extracted from depth. With the recent development of deep learning, a few methods~\cite{pichao2015,pichaoTHMS} have been developed based on Convolutional Neural Networks (ConvNets). A common and intuitive method to represent human motion is to use a sequence of skeletons. With the development of the cost-effective depth cameras and algorithms for real-time pose estimation~\cite{Shotton2011}, skeleton extraction has become more robust and many hand-designed skeleton features~\cite{Yang2012,zanfir2013moving,Gowayyed2013_HOD,pichao2014,vemulapalli2014human} for action recognition have been proposed. Recently, Recurrent Neural Networks (RNNs)~\cite{du2015hierarchical,veeriah2015differential,zhu2015co,shahroudy2016ntu} have also been adopted for action recognition from skeleton data. The hand-crafted features are always shallow and dataset-dependent. RNNs tend to overemphasize the temporal information especially when the training data is not sufficient, leading to overfitting. In this paper, we present a compact, effective yet simple method that encodes the joint trajectories into texture images, referred to as Joint Trajectory Maps (JTM), as the input of ConvNets for action recognition. In this way, the capability of the ConvNets in learning discriminative features  can be fully exploited~\cite{zeiler2014visualizing}. 

One of the challenges in action recognition is how to properly model and use the spatio-temporal information. The commonly used bag-of-words model tends to overemphasize the spatial information. On the other hand, Hidden Markov Model (HMM) or RNN based methods are likely to overstress the temporal information. The proposed method addresses this challenge in a different way by encoding as much the spatio-temporal information as possible (without a need to decide which one is important and how important it is) into images and letting the CNNs to learn the discriminative one. This is the key reason that the proposed method outperformed previous ones. In addition, the proposed encoding method can be extended to online recognition due to the accumulative nature of the encoding process. Furthermore, such encoding of spatio-temporal information into images allows us to leverage the advanced methods developed for image recognition.

\section{The Proposed Method}

\begin{figure*}[!ht]
\begin{center}
{\includegraphics[height = 60mm, width = 160mm]{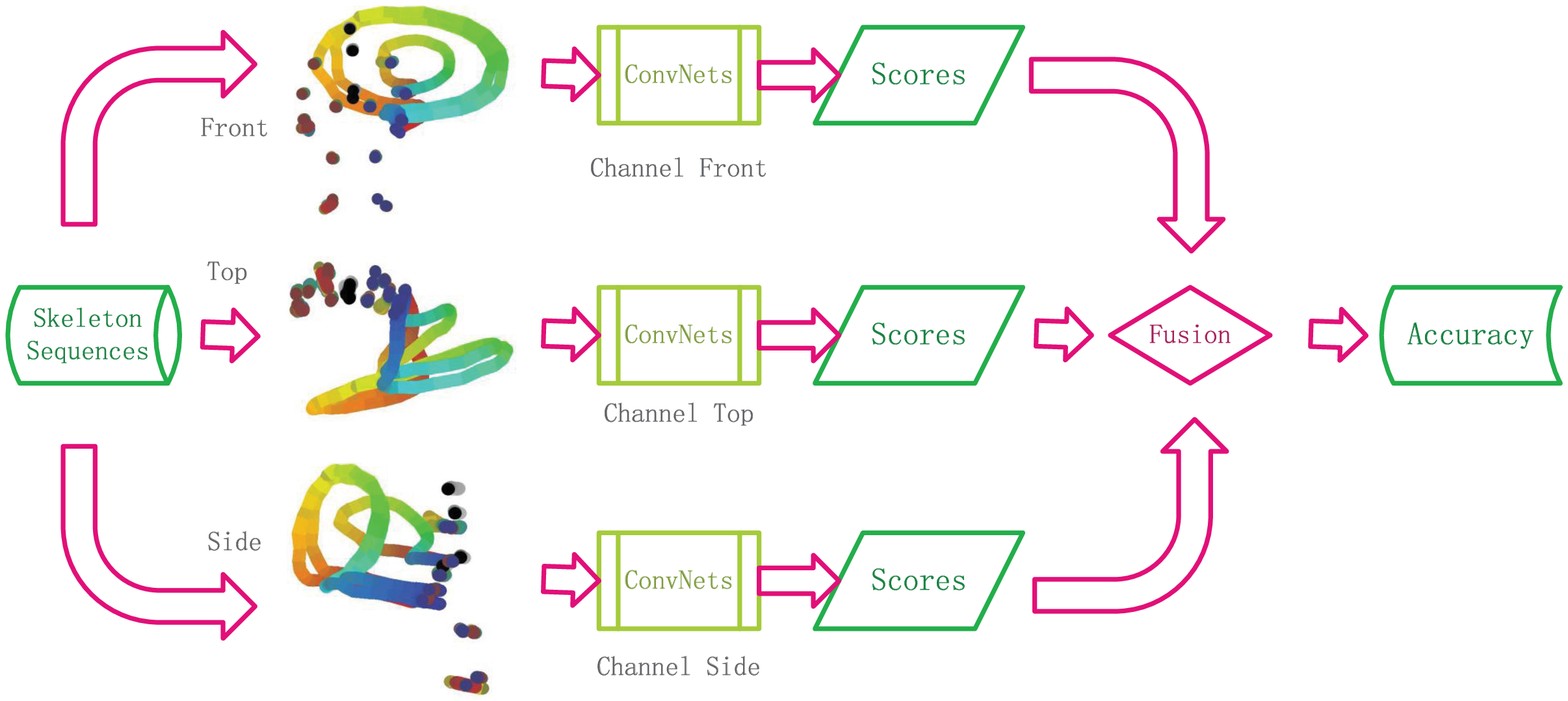}}
\end{center}
\caption{The framework of the proposed method.}
\label{fig:framework}
\end{figure*}

The proposed method consists of two major components, as illustrated in Fig.~\ref{fig:framework}, three ConvNets and the construction of three JTMs as the input of the ConvNets in three orthogonal planes from the skeleton sequences. Final classification of a given test skeleton sequence is obtained through a late fusion of the three ConvNets. The main contribution of this paper is on the construction of suitable JTMs for the ConvNets to learn discriminative features.

We argue that an effective JTM should have the following properties to keep sufficient spatial-temporal information of an action:
\begin{itemize}
\item The joints or group of joints should be distinct in the JTM such that the spatial information of the joints is well reserved.
\item The JTM should encode effectively the temporal evolution, i.e. trajectories of the joints, including the direction and speed of joint motions.
\item The JTM should be able to encode the difference in motion among the different joints or parts of the body to reflect how the joints are synchronized during the action.
\end{itemize}

Specifically, JTM can be recursively defined  as follows
\begin{equation}
JTM_{i} = JTM_{i -1} + f(i)
\label{eq:spectrum}
\end{equation}
where $f(i)$ is a function encoding the spatial-temporal information at frame or time-stamp $i$. Since JTM is accumulated over the period of an action, $f(i)$ has to be carefully defined such that the JTM for an action sample has the required properties and the accumulation over time has little adverse impact on the spatial-temporal information encoded in the JTM. 
We propose in this paper to use hue, saturation and brightness to encode the spatial-temporal motion patterns.

\subsection{Joint Trajectory Maps}
Assume an action $H$ has $n$ frames of skeletons and each skeleton consists of $m$ joints. The skeleton sequence is denoted as $H=\{F_{1},F_{2},...,F_{n}\}$, where $F_{i} = \{P_{1}^{i},P_{2}^{i},...,P_{m}^{i}\}$ is a vector of the joint coordinates at frame $i$, and $P_{j}^{i}$ is the $3D$ coordinates of the $j$th joint in frame $i$. The skeleton trajectory $T$ for an action of $n$ frames consists of the trajectories of all joints and is defined as:
\begin{equation}
T = \{T_1, T_2, \cdots,T_i,\cdots,T_{n-1}\}
\label{eq:actionrepre}
\end{equation}
where $T_{i} = \{t_{1}^{i},t_{2}^{i},...,t_{m}^{i}\} = F_{i+1} - F_{i}$ and the $k$th joint trajectory is $t_{k}^{i} = P_{k}^{i+1} - P_{k}^{i}$. At this stage, the function $f(i)$ is the same as $T_{i}$, that is,
\begin{equation}
f(i) = {T_i} = \{ {t_1^i,t_2^i,...,t_m^i} \}.
\end{equation}
 
\begin{figure}[!ht]
\begin{center}
{\includegraphics[height = 30mm, width = 85mm]{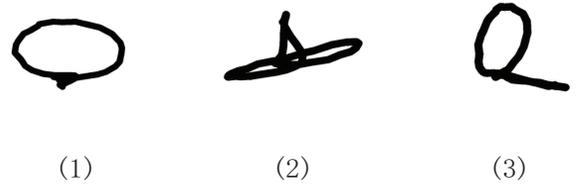}}
\end{center}
\caption{The trajectories projected onto three Cartesian planes for action ``right hand draw circle (clockwise)" in {UTD-MHAD~\protect\cite{chenchen2015b}}: (1) the front plane; (2) the top plane; (3) the side plane.}
\label{fig1}
\end{figure}

The skeleton trajectory is projected to the three orthogonal planes, i.e. three Cartesian planes, to form three JTMs. Fig.~\ref{fig1} shows the three projected trajectories of the right hand joint for action ``right hand draw circle (clockwise)" in the UTD-MHAD dataset. From these JTMs, it can be seen that the spatial information of this joint is preserved but the direction of the motion is lost.

\subsection{Encoding Joint Motion Direction} 
\begin{figure}[!ht]
\begin{center}
{\includegraphics[height = 40mm, width = 85mm]{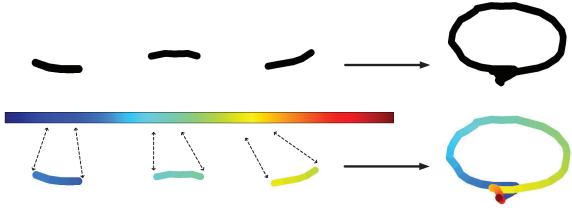}}
\end{center}
\caption{The trajectories of different body parts have their different colors reflecting the temporal orders.}
\label{fig2}
\end{figure}

To capture the motion information in the JTM, it is proposed to use hue to represent the motion direction. Different kinds of colormaps can be chosen. In this paper, the jet colormap, ranging from blue to red, and passing through the colors cyan, yellow, and orange, was adopted. Assume the color of a joint trajectory is $C$ and the length of the trajectory $L$, and let $C_{l}, l \in (0, L)$ be the color at position $l$. For the $q^{th}$ trajectory $T_{q}$ from $1$ to $n-1$, a color $C_{l}$, where $l = \frac{q}{n-1}\times L$ is specified to the joint trajectory, making different trajectories have their own color corresponding to their temporal positions in the sequence as illustrated in Fig.~\ref{fig2}. Herein, the trajectory with color is denoted as $C\_t_k^i$ and the function $f(i)$ is updated to:
\begin{equation}
f(i) = \{ {C\_t_1^i,C\_t_2^i,...,C\_t_m^i} \}.
\end{equation}
This ensures that different actions are encoded to a same length colormap. The effects can be seen in Fig.~\ref{fig3}, sub-figures (1) to (2). Even though the same actions with different number of cycles will be encoded into different color shapes, the direction can still be reflected in color variation and the differences between actions can still be captured due to the different spatial information.

\subsection{Encoding Body Parts}
To distinguish different body parts, multiple colormaps are employed.
There are many ways to achieve this. For example, each joint is assigned to one colormap, or several groups of joints are assigned to different colormaps randomly. Considering arms and legs often have more motion than other body parts, we empirically generate three colormaps ($C1, C2, C3$) to encode three body parts. $C1$ is used for the left body part (consisting of left shoulder, left elbow, left wrist, left hand, left hip, left knee, left ankle and left foot), $C2$ for the right body part ( consisting of right shoulder, right elbow, right wrist, right hand, right hip, right knee, right ankle and right foot), and $C3$ for the middle body part (consisting of head, neck, torso and hip center). $C1$ is the same as $C$, i.e. the jet colormap, $C2$ is a reversed colormap of $C1$, and $C3$ is a colormap ranging from light gray to black. Here, the trajectory encoded by multiple colormaps is denoted as $MC\_t_k^i$, and the function $f(i)$ is formulated as:
\begin{equation}
f(i) = \{ {MC\_t_1^i,MC\_t_2^i,...,MC\_t_m^i} \}.
\end{equation}
 The effects can be seen in Fig.~\ref{fig3}, sub-figures (2) to (3).

\subsection{Encoding Motion Magnitude}
\begin{figure}[!ht]
\begin{center}
{\includegraphics[height = 60mm, width = 80mm]{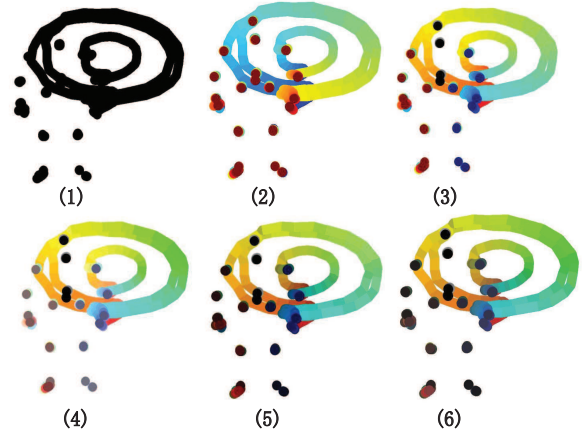}}
\end{center}
\caption{Illustration of visual differences for different techniques in JTM.}
\label{fig3}
\end{figure}

Motion magnitude is one of the most important factors in human motion. For one action, large magnitude of motion usually indicates more motion information. In this paper, it is proposed to encode the motion magnitude of joints into the saturation and brightness components, so that such encoding not only encodes the motion but also enriches the texture of trajectories which are expected to be beneficial for ConvNets to learn discriminative features. For joints with high motion magnitude or speed, high saturation will be assigned as high motion usually carries more discriminative information. Specifically, the saturation is set to range from $s_{min}$ to $s_{max}$.  Given a trajectory, its saturation $S_{j}^{i}$ in $HSV$ color space could be calculated as
\begin{equation}
S_{j}^{i} = \frac{v_{j}^{i}}{max\{v\}} \times (s_{max} - s_{min}) + s_{min}
\label{eq:velocity}
\end{equation}
where $v_{j}^{i}$ is the $j$th joint speed at the $i$th frame.
\begin{equation}
v_{j}^{i} = \Vert P_{j}^{i+1} - P_{j}^{i} \Vert_{2}
\label{eq:velocity}
\end{equation}
The trajectory adjusted by saturation is denoted as $M{C_s}\_t_k^i$ and the function $f(i)$ is refined as:
\begin{equation}
f(i) = \{{M{C_s}\_t_1^i,M{C_s}\_t_2^i,...,M{C_s}\_t_m^i} \}
\end{equation}
The encoding effect can be seen in Figure~\ref{fig3}, sub-figures (3) to (4), where the slow motion becomes diluted (e.g. trajectory of knees and ankles) while the fast motion becomes saturated (e.g. the green part of the circle).

To further enhance the motion patterns in the JTM, the brightness is modulated by the speed of joints so that motion information is enhance in the JTM by rapidly changing the brightness according to the joint speed. In particular, the brightness is set to range from $b_{min}$ to $b_{max}$. Given a trajectory $t_{j}^{i}$ whose speed is $v_{j}^{i}$, its brightness $B_{j}^{i}$ in the $HSV$ color space is calculated as
\begin{equation}
B_{j}^{i} = \frac{v_{j}^{i}}{max\{v\}} \times (b_{max} - b_{min}) + b_{min}
\label{eq:velocity1}
\end{equation}
The trajectory adjusted by brightness is denoted as $M{C_b}\_t_k^i$ and the function $f(i)$ is updated to:
\begin{equation}
f(i) = \{ {M{C_b}\_t_1^i,M{C_b}\_t_2^i,...,M{C_b}\_t_m^i} \}. 
\end{equation}
The effect can be seen in Fig~\ref{fig3}, sub-figures (3) to (5), where texture becomes apparent (e.g. the yellow parts of the circle).
Finally, motion magnitude is encoded with saturation and brightness together. The trajectory is denoted as $M{C_{sb}}\_t_k^i$ and the function $f(i)$ is refined as:
\begin{equation}
f(i) = \{ {M{C_{sb}}\_t_1^i,M{C_{sb}}\_t_2^i,...,M{C_{sb}}\_t_m^i} \}.
\end{equation}
 As illustrated in Fig.~\ref{fig3}, sub-figures(3) to (6), it not only enriches the texture information but also highlights the faster motion.

\subsection{Training and Recognition}
In the experiments, the layer configuration of the three ConvNets was same as the one in~\cite{krizhevsky2012imagenet}. The implementation was derived from the publicly available Caffe toolbox \cite{jia2014caffe} based on one {NVIDIA GeForce GTX TITAN X} card and the pre-trained models over ImageNet~\cite{krizhevsky2012imagenet} were used for initialization in training. The network weights are learned using the mini-batch stochastic gradient descent with the momentum being set to 0.9 and weight decay being set to 0.0005. At each iteration, a mini-batch of 256 samples is constructed by sampling 256 shuffled training JTMs. All JTMs are resized to 256 $\times$ 256. The learning rate is to $10^{-3}$ for fine-tuning and then it is decreased according to a fixed schedule, which is kept the same for all training sets. For each ConvNet the training undergoes 100 cycles and the learning rate decreases every 20 cycles. For all experiments, the dropout regularisation ratio was set to 0.5 in order to reduce complex co-adaptations of neurons in nets. Three ConvNets are trained on the JTMs in the three Cartesian planes and the final score for a test sample are the averages of the outputs from the three ConvNets. The testing process can easily achieved real-time speed (average 0.36 seconds/sample).

\section{Experimental Results}
The proposed method was evaluated on three public benchmark datasets: MSRC-12 Kinect Gesture Dataset~\cite{Fothergill2011}, G3D~\cite{bloom2012g3d} and UTD-MHAD~\cite{chenchen2015b}. Experiments were conducted to evaluate the effectiveness of each encoding scheme in the proposed method and the final results were compared with the state-of-the-art reported on the same datasets. In all experiments, the saturation and brightness covers the full range (from 0\% $\sim$ 100\% mapped to 0 $\sim$ 255) in HSV color space.

\subsection{Evaluation of Different Encoding Schemes}

The effectiveness of different encoding schemes (corresponding to the sub-figures in \ref{fig3}) was evaluated on the G3D dataset using the front JTM and the recognition accuracies are listed in Table~\ref{steps}.

\begin{table}[!th]
\centering

\begin{tabular}{|c|c|}
\hline
Techniques & Accuracy (\%)\\
\hline
Trajectory: $t_1^i$ & 63.64\%\\
\hline
Trajectory: $C\_t_1^i$ & 74.24\%\\
\hline
Trajectory: $MC\_t_1^i$ & 78.48\%\\
\hline
Trajectory: $M{C_s}\_t_1^i$ & 81.82\%\\
\hline
Trajectory: $M{C_b}\_t_1^i$ & 82.12\%  \\
\hline
Trajectory: $M{C_{sb}}\_t_1^i$ & 85.45\%  \\
\hline
\end{tabular}
\caption{Comparisons of the different encoding schemes on the G3D dataset using the JTM projected to the front plane alone.\label{steps}}
\end{table}
From this Table it can be seen that the proposed encoding techniques effectively captures the spatio-temporal information and the ConvNets are able to learn the discriminative features from the JTM for action recognition.


\subsection{MSRC-12 Kinect Gesture Dataset}
MSRC-12~\cite{Fothergill2011} is a relatively large dataset for gesture/action recognition from 3D skeleton data captured by a Kinect sensor. The dataset has 594 sequences, containing 12 gestures by 30 subjects, 6244  gesture instances in total. The 12 gestures are: ``lift outstretched arms", ``duck", ``push right", ``goggles", ``wind it up", ``shoot",  ``bow", ``throw", ``had enough", ``beat both", ''change weapon" and ``kick". For this dataset, cross-subjects protocol is adopted, that is odd subjects for training and even subjects for testing. Table~\ref{table1} lists the performance of the proposed method and the results reported before.

\begin{table}[!th]
\centering

\begin{tabular}{|c|c|}
\hline
Method & Accuracy (\%)\\
\hline
HGM~\cite{yang2014hierarchical} & 66.25\%\\
\hline
ELC-KSVD~\cite{zhou2014discriminative} & 90.22\%\\
\hline
Cov3DJ~\cite{Hussein2013} & 91.70\%\\
\hline
Proposed Method & \textbf{93.12\%}  \\
\hline
\end{tabular}
\caption{Comparison of the proposed method with the existing methods on the MSRC-12 Kinect gesture dataset.\label{table1}}
\end{table}

\begin{figure}[!ht]
\begin{center}
{\includegraphics[height = 50mm, width = 85mm]{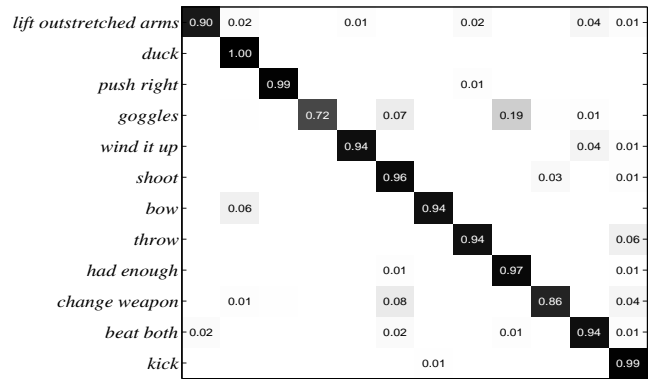}}
\end{center}
\caption{The confusion matrix of the proposed method for MSRC-12 Kinect gesture dataset.}
\label{fig:confusion1}
\end{figure}

The confusion matrix is shown in figure~\ref{fig:confusion1}. From the confusion matrix we can see that the proposed method distinguishes most of actions very well, but it is not very effective to distinguish ``goggles" and ``had enough" which shares the similar appearance of JTM probably caused by 3D to 2D projection.

\subsection{G3D Dataset}
Gaming 3D Dataset (G3D)~\cite{bloom2012g3d} focuses on real-time action recognition in gaming scenario. It contains 10 subjects performing 20 gaming actions: ``punch right", ``punch left", ``kick right", ``kick left", ``defend", ``golf swing", ``tennis swing forehand", ``tennis swing backhand", ``tennis serve", ``throw bowling ball", ``aim and fire gun", ``walk", ``run", ``jump", ``climb", ``crouch", ``steer a car", ``wave", ``flap" and ``clap".
For this dataset, the first 4 subjects were used for training, the fifth for validation and the remaining 5 subjects for testing as configured in~\cite{Nie2015}.

Table~\ref{table2} compared the performance of the proposed method and that reported in~\cite{Nie2015}.

\begin{table}[ht!]
\centering
\begin{tabular}{|c|c|}
\hline
Method & Accuracy (\%)\\
\hline
LRBM \cite{Nie2015} & 90.50\%  \\
\hline
Proposed Method & \textbf{94.24\%}\\
\hline
\end{tabular}
\caption{Comparison of the proposed method with previous methods on G3D Dataset.\label{table2}}
\end{table}

\begin{figure}[!ht]
\begin{center}
{\includegraphics[height = 50mm, width = 85mm]{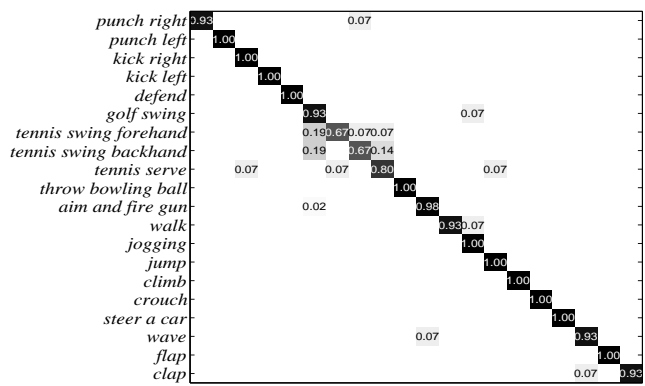}}
\end{center}
\caption{The confusion matrix of the proposed method for G3D Dataset.}
\label{fig:confusion2}
\end{figure}

The confusion matrix is shown in figure~\ref{fig:confusion2}. From the confusion matrix we can see that the proposed method recognizes most of actions well. Compared with LRBM, our proposed method outperforms LRBM in spatial information mining. LRBM confused mostly the actions between ``tennis swing forehand" and ``bowling", ``golf" and ``tennis swing backhand", ``aim and fire gun" and ``wave", ``jump" and ``walk", however, these actions were quite well distinguished in our method because of the good spatial information exploitation in our method. As for ``aim and fire gun" and ``wave", our method could not distinguish them well before encoding the motion magnitude, which means the temporal information enhancement procedure is effective. However, in our method, ``tennis swing forehand" and ``tennis swing backhand" are confused. It's probably because the front and side projections of body shape of the two actions are too similar, and scores fusion is not very effective to improve each other.

\subsection{UTD-MHAD}

UTD-MHAD~\cite{chenchen2015b} is one multimodal action dataset, captured by one Microsoft Kinect camera and one wearable inertial sensor. This dataset contains 27 actions performed by 8 subjects (4 females and 4 males) with each subject perform each action 4 times. After removing three corrupted sequences, the dataset includes 861 sequences. The actions are: ``right arm swipe to the left", ``right arm swipe to the right", ``right hand wave", ``two hand front clap", ``right arm throw", ``cross arms in the chest", ``basketball shoot", ``right hand draw x", ``right hand draw circle (clockwise)", ``right hand draw circle (counter clockwise)", ``draw triangle", ``bowling (right hand)", ``front boxing", ``baseball swing from right", ``tennis right hand forehand swing", ``arm curl (two arms)", ``tennis serve", ``two hand push", ``right hand know on door", ``right hand catch an object", ``right hand pick up and throw", ``jogging in place", ``walking in place", ``sit to stand", ``stand to sit", ``forward lunge (left foot forward)" and ``squat (two arms stretch out)". It covers sport actions (e.g. ``bowling", ``tennis serve" and ``baseball swing"), hand gestures (e.g. ``draw X", ``draw triangle", and ``draw circle"), daily activities (e.g. ``knock on door", ``sit to stand" and ``stand to sit") and training exercises (e.g. ``arm curl", ``lung" and ``squat"). For this dataset, cross-subjects protocol is adopted as in ~\cite{chenchen2015b}, namely, the
data from the subject numbers 1, 3, 5, 7 used for training while 2, 4, 6, 8 
used for testing.

Table~\ref{table3} compared the performance of the proposed method and that reported in~\cite{chenchen2015b}.

\begin{table}[ht!]
\centering
\begin{tabular}{|c|c|}
\hline
Method & Accuracy (\%)\\
\hline
Kinect \& Inertial \cite{chenchen2015b} & 79.10\%  \\
\hline
Proposed Method & \textbf{85.81\%}\\
\hline
\end{tabular}
\caption{Comparison of the proposed method with previous methods on UTD-MHAD Dataset.\label{table3}}
\end{table}
Please notice that the method used in~\cite{chenchen2015b} is based on Depth and Inertial sensor data, not skeleton data alone.

\begin{figure}[!ht]
\begin{center}
{\includegraphics[height = 50mm, width = 85mm]{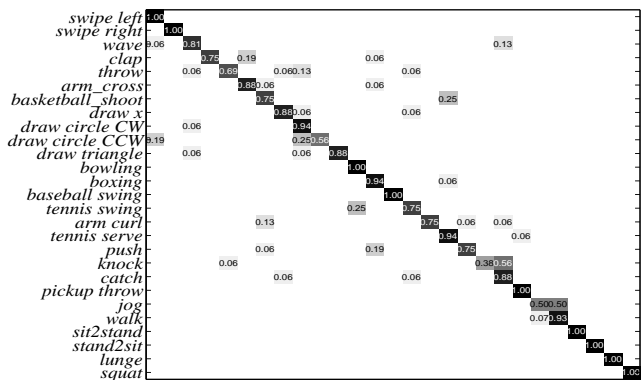}}
\end{center}
\caption{The confusion matrix of the proposed method for UTD-MHAD.}
\label{fig:confusion3}
\end{figure}

The confusion matrix is shown in figure~\ref{fig:confusion3}. This dataset is much more challenging compared to previous two datasets. From the confusion matrix we can see that the proposed method can not distinguish some actions well, for example, ``jog" and ``walk". A probable reason is that the proposed encoding process is also a normalization process along temporal axis (Section 3.2). The actions ``jog" and ``walk" will be normalized to have a very similar JTM after the encoding.

\section{Conclusion}
This paper addressed the problem of human action recognition by applying ConvNets to skeleton sequences. We proposed an effective method to encode the joints trajectories to JTM where the motion information can be encoded into texture patterns. ConvNets learn discriminative features from these maps for real-time human action recognition. The experimental results showed that the techniques for encoding worked effectively. The proposed method can benefit from effective data augmentation process which would be our future work.

\section{Acknowledgments}
This work was supported by the National Natural Science Foundation of China (grant 61571325) and Key Projects in the Tianjin Science \& Technology Pillar Program (grant 15ZCZD GX001900).

\end{document}